\documentclass[reprint,amsmath,amssymb,aps,floatfix]{revtex4-2}

\usepackage{graphicx}
\usepackage{dcolumn}
\usepackage{bm}%
\usepackage{xcolor}
\usepackage{subcaption}
\usepackage{caption}
\usepackage{hyperref}
\usepackage{url}
\begin{document}

\preprint{APS/123-QED}

\title{First Experimental Demonstration of Natural Hovering Extremum Seeking: A New Paradigm in Flapping Flight Physics}

\author{Simone Martini}
\affiliation{Department of Aerospace Engineering and Engineering Mechanics, 
University of Cincinnati, Cincinnati, Ohio 45221, USA}

\author{Rohan~Palanikumar}
\affiliation{Department of Aerospace Engineering and Engineering Mechanics, 
University of Cincinnati, Cincinnati, Ohio 45221, USA}

\author{Ahmed A.~Elgohary}
\affiliation{Department of Aerospace Engineering and Engineering Mechanics, 
University of Cincinnati, Cincinnati, Ohio 45221, USA}

\author{Sameh A.~Eisa}
\affiliation{Department of Aerospace Engineering and Engineering Mechanics, 
University of Cincinnati, Cincinnati, Ohio 45221, USA}

\begin{abstract}
In this article, we report the first attempt of experimental validation of the recently emerged new
paradigm in hovering and flapping flight physics called ``Natural Hovering Extremum Seeking (NH-ES)" [doi.org/10.1103/4dm4-kc4g], which theorized that stable hovering flight physics observed in nature by flapping insects and hummingbirds can be characterized and emulated via a model-free, real-time, computationally-basic, sensory-based feedback mechanism that only needs the built-in natural oscillations of the flapping wing motion as both the control and the propulsion input. We run experiments of \textit{moth-like, light bulb source-seeking}, on a flapping-wing body in a total model-free setting that is agnostic to morphological parameters and body/aerodynamic models. {That is, the flapping performer (e.g., insects or hummingbirds) do not perceive internal or external governing physical equations, system states, models or parameters. Moreover, in addition to experimentally demonstrating the NH-ES paradigm in vertical motion as predicted by simulations in our earlier foundational work, we provide experiments that suggest the expansion of the NH-ES theory to longitudinal mode by yielding altitude and pitch stabilization in hovering flight. In the presented experiments, the flapper \textit{seeks, then hovers, around the light bulb source like a moth}}, needing only feedback of sensory measurements of light intensity {taken by a simple sensor mounted on the head of the flapper body. Our experiment confirms that NH-ES needs only instantaneous, local measurement of the physical signal (e.g., light) as a feedback to perform source-seeking and does not need measurements, estimations or reference information of states, angles, altitude, or coordinates of the flapper system or source, nor does it need any global information such as access to GPS.} Our results were also achieved under delay and noise effects, supporting earlier observations that NH-ES is robust against potential processing delays and noisy-sensations.
\end{abstract}

\maketitle
\section{Introduction}
Flapping flight is a natural physical phenomenon that has captivated research interest across many scientific and engineering fields, including physics, biology, computational mathematics, and aerospace, control, and robotic engineering \cite{sun2014insect,xuan2020recent,taha_review,phan2019insect,hedrick2015recent}. Physical modeling of flapping flight has long been recognized as a highly complex problem, involving nonlinear multibody, multiple-time-scale, and non-autonomous dynamics \cite{liang2013nonlinear,taha_review,IEEETransaction,phan2019insect,maggia2020higher}. Among flapping maneuvers, \emph{hovering} is particularly intriguing because it demands lift coefficients far beyond those predicted by conventional aerodynamic principles developed for fixed-wing aircraft \cite{weis1972energetics,ellington1995unsteady}. Over time, scientists and engineers have identified unconventional lift mechanisms enabling hovering flight, such as the leading-edge vortex \cite{nature1996LEV,dickinson1999wing}, which accounts for the necessary lift in insect flight. While this body of literature has succeeded in answering fundamental questions about the physics of hovering, it remains to explain the mechanism used in nature by flapping insects and hummingbirds to achieve such complex physics in a stable and real-time manner. {In other words, what is the control mechanism used by insects and hummingbirds that enable and explain hovering midair stably in real-time?}

The long-held consensus in literature is that hovering flapping flight is inherently unstable in the open-loop sense for most insects, and that an active control mechanism is therefore required for stabilization \cite{taha_review,sun2014insect,IEEETransaction,deng2006flapping,taylor2003dynamic,taylor2005nonlinear,feedback25ristroph2010discovering,wang2000two,lyu2022dynamic,taha2020vibrational}; see recent investigation on this topic \cite{abdelfattah2025optimal}. At the same time, biological observations and experiments consistently indicate that flapping insects rely heavily on sensory feedback for stable flight and \textit{stabilized hovering} \cite{fuller2014flying,sensorroyalristroph2013active,taylor2003dynamic,feedback25ristroph2010discovering,sensors1taylor2007sensory}. These observations suggest that the active control mechanism employed in nature is sensation-based and operates under limited computational capabilities \cite{nature_toaster,taha2020vibrational,deng2006flapping}. Consequently, the case was made in \cite{elgohary2025hovering} that a sensation-based control mechanism for hovering {is biologically plausible if it is simple, computationally-basic and independent of explicit aerodynamic or body perception (i.e., model-free).} {The reader can refer to \cite[Section I]{elgohary2025hovering} for more on control methods used in hovering/flapping flight literature}. {We note to the reader that in a later section we will recall some of these control methods to highlight the categorical difference and biological plausibility of our new paradigm.}  

Very recently, we theorized and proposed a new paradigm in hovering and flapping flight physics, which we call \emph{Natural Hovering Extremum Seeking (NH-ES) \cite{elgohary2025hovering}}. The concept of NH-ES is grounded in a recent mathematical control framework \cite{elgohary2025extremum} {called Extremum Seeking for Vibrational Stabilization (ES-VS)} that takes advantage of a high-amplitude, high-frequency input signal to steer a physical system towards stabilization using simple sensory feedback, which operates as a continuously adjusted vibrational stabilization mechanism. {Maintaining the same appeal of previous extremum seeking methods (see for example \cite{tan2010extremum,ariyur2003real,pokhrel2023higher,eisa2023analyzing,scheinker2024100,grushkovskaya2018class}), ES-VS idea is to actuate perturbations into a dynamical system, then via access to corresponding variations in the objective function measurements, a simple feedback estimate loop is used to adjust the input to steer the dynamical system to the vicinity of the extremum point (or optimal state) of the objective function. In essence, extremum seeking methods are an autonomous, systematic and stable ``trial and error", ``perturb and observe" or {``online search and adapt"} control mechanism as they are operable without access to the dynamic system model or the objective function formula; that is, one only needs to actuate the perturbations and measure the objective function. However, ES-VS \cite{elgohary2025extremum,palanikumar2026model} is distinguished from previous extremum seeking methods \cite{tan2010extremum,ariyur2003real,pokhrel2023higher,eisa2023analyzing,scheinker2024100,grushkovskaya2018class} in that it needs \textit{only one} high-amplitude, high frequency perturbation signal to vibrationally stabilize and control a second-order mechanical system that can admit quadratic forces in velocities, allowing the inclusion of aerodynamic forces {and direct treatment of the system in its second-order ``equation-of-motion" governed physics.} As a result, ES-VS made it possible to characterize hovering flight physics as NH-ES \cite{elgohary2025hovering} with the \text{only one} high amplitude, high frequency perturbation signal being the flapping motion itself, naturally built in the system for propulsion! The reader is directed to the supplementary file ``Tutorial.pdf" (also available here \cite{github_flapping_experiment}) for a step-by-step tutorial on how ES-VS concept works using simple 1D inverted pendulum and mass-spring examples for demonstration.}

Since it was theorized in \cite{elgohary2025hovering}, NH-ES attracted significant scientific attention {(see for example \cite{wilkinson2025flapping,PhysicsWorld} and the statistics published by the publisher \cite{Attention})}. In \cite{elgohary2025hovering}, NH-ES was successfully validated {via: (i) positive stability analysis using sophisticated tools from differential geometric control theory \cite[Section IV]{elgohary2025hovering}; (ii) effective and robust} simulations across multiple insect species and a hummingbird data set {\cite[Section III.A]{elgohary2025hovering}} {which closely matched reported biological data \cite[Table IV]{elgohary2025hovering}}, all of which by using the same feedback mechanism and learning-rate parameter {\cite[Table III]{elgohary2025hovering}}; {(iii) its ability to tolerate processing delays and sensation noise \cite[Section III.D]{elgohary2025hovering}. 
Given that the NH-ES feedback mechanism is sensation-based, model-free, real-time, closely matched with biological data, computationally-minimal, {its input is the already built-in propulsion mechanism of flapping wing motion}, it has been theorized as biologically plausible. {Nevertheless, the new paradigm, theory, and concept of NH-ES \cite{elgohary2025hovering} was not tested by simulations that include pitch dynamics (a main reason for instability in open-loop hovering, especially for small insects \cite{liang2013nonlinear,taha2020vibrational}), nor was it experimentally verified. Needless to say, the new theory and paradigm of NH-ES needs to take the path and traditions followed in natural physical sciences and move into applied physics verification}.

In this {paper,} we present the first attempt of experimental demonstration of NH-ES, aiming to verify, {and open the path for}, this paradigm-shifting concept in hovering and flapping flight physics. {Our aim is to: (1) validate the initial simulations of purely vertical-motion flapping, findings/observations in \cite{elgohary2025hovering}; and (2) demonstrate the ability of the NH-ES to go beyond vertical-motion flapping and include pitching dynamics (i.e., longitudinal mode). 
{We run experiments of \textit{moth-like, light bulb source-seeking}, on a flapping-wing body in a total model-free setting that is agnostic to morphological parameters and body/aerodynamic models. Videos for each experiment are provided as supplementary materials and also available in \cite{MDCL_youtube}. {That is, the flapping performer (e.g., insects or hummingbirds) do not need to perceive internal or external governing physical equations, models or parameters.} We show that the flapper using NH-ES gains altitude and stabilizes {autonomously the servos responsible for flapping, \textit{including pitching dynamics}. The flapper effectively and stably} \textit{hovers around the light bulb source like a moth}, needing only feedback of sensory measurements of light intensity {taken by a simple sensor mounted on the head of the flapper body, confirming that NH-ES needs only instantaneous, local measurement of the physical signal (e.g., light) as feedback and does not need measurements, estimations or reference information of states, angles, altitude, or coordinates of the flapper system or the light source, nor does it need any global information such as access to GPS}. These characteristics place NH-ES as a complete new paradigm which is categorically different from available control mechanisms in literature (discussed more in Section V.A).} 
Furthermore, our results are:
\begin{itemize}
    \item able to confirm that the high-amplitude, high-frequency flapping wing motion results in variations in sensory measurements on the body. This confirms the main NH-ES conceptual premise. 
    \item conducted successfully in real-time under inherent processing delays and measurement noises. Hence, they confirm the autonomous operability of NH-ES and its robustness against noise and delays.
    \item successful in demonstrating stable source-seeking to, then hovering about, a desired altitude mimicking a ``moth-like" light source-seeking and hovering. The NH-ES design shows stable actuation of the servos responsible for flapping, and pitch dynamics.  
\end{itemize}

\section{Natural Hovering Extremum Seeking Design and Experimental Setup}
\begin{figure}[t]
    \centering
    \includegraphics[width=\linewidth]{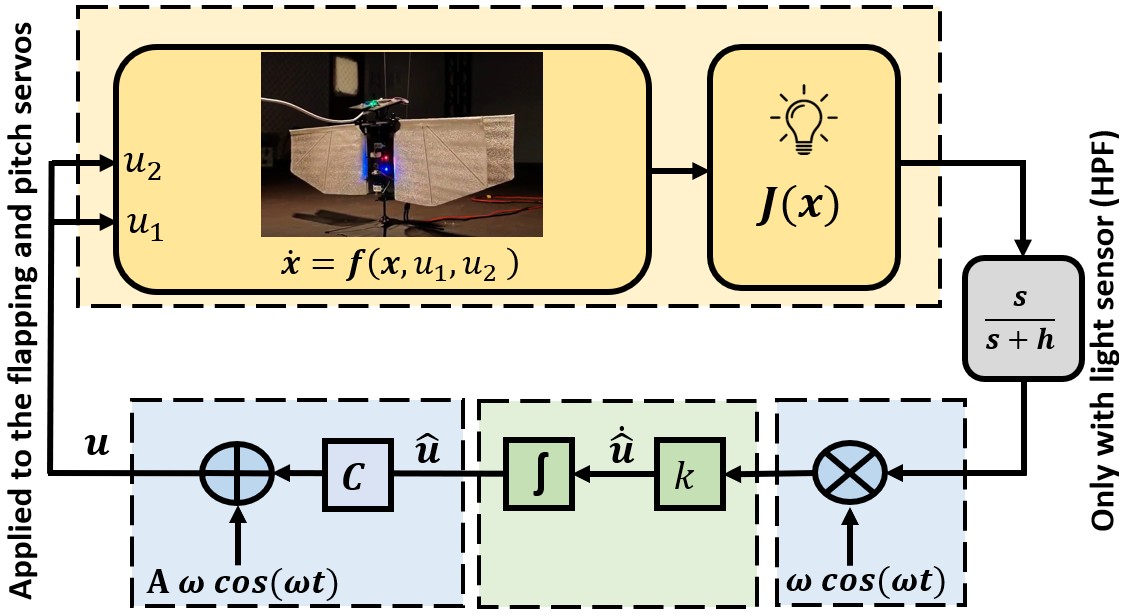}
    \caption{{Model-free Natural Hovering Extremum Seeking (NH-ES) control architecture; more details and specifics are provided in ``Setup.pdf". The flapping-wing system dynamics $f(\cdot)$ and the objective function $J(\cdot)$ are treated as black boxes. The control input is written as the vector $\boldsymbol{u}=[u_1\;u_2]^T$, where $u_1$ represents the flapping motor input and $u_2$ represents the pitch servo input. The adaptive component $\hat{u}$ is combined with the oscillatory term naturally generated by wing flapping, $k$ is the learning parameter, and $C$ is a gain parameter. The high pass filter (HPF) is optional for sensory measurements for potential refinement {and is consistent with common practice in experimental implementations in literature \cite{bajpai2024model,elgohary2025model2,palanikumar2026model2}}.}}
    \label{fig:nhes_block}
\end{figure}
\begin{figure}[]
    \centering
    \includegraphics[width=\linewidth]{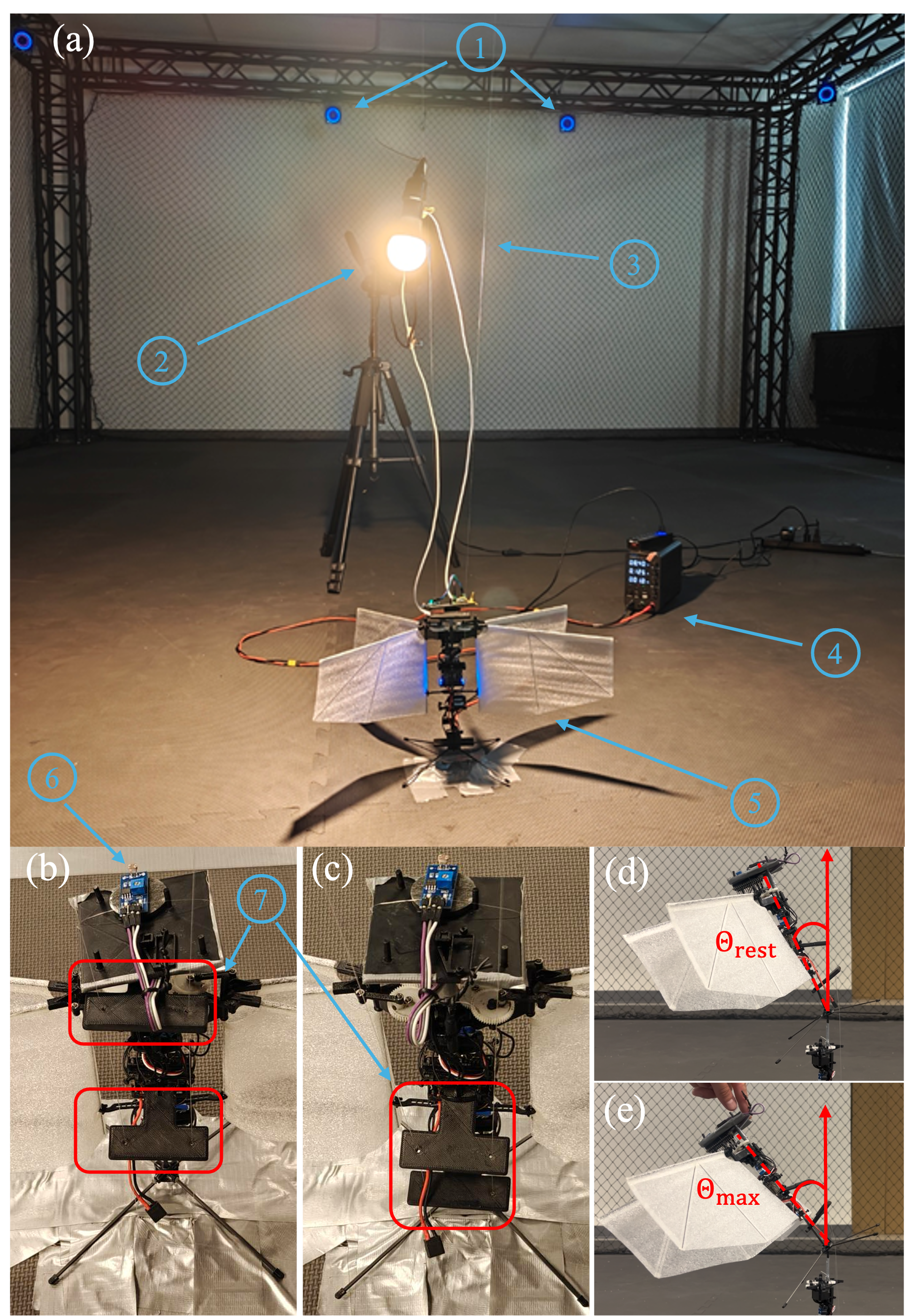}
    \caption{Experimental setup for {moth-like, light source seeking by a flapping body system} relying exclusively on intrinsic flapping-induced oscillations and scalar sensory feedback. {In Section III experiments, we focus on vertical motion and relatively fixed pitch, while experiments in Section IV include both vertical and pitching dynamics.}
    (\#1): Motion-capturing system (MCS). (\#2): Light source. (\#3): Two constraint lines with relaxed tension. (\#4): External power supply. (\#5): Flapping-wing body. {(\#6): Light Sensor. (\#7): Line-Guides (fixed on the flapping body for the constraint lines to go through them). (a) Displays the overall experimental setup. (b) Shows the Line-Guides positioned far apart (one at the top by the head and one near the middle height) to minimize pitching motion and isolate the vertical dynamics for Section III experiments. (c) Shows the Line-Guides positioned close apart (near the middle height) to allow pitching motion for the Section IV experiments. {(d) and (e) Display the maximum resting pitch angle ($\theta_{\text{rest}} \approx 25$ deg) and maximum pitch angle ($\theta_{\text{max}} \approx 40$ deg) for Line-Guides setup (c), obtained without and with external force applied to the flapping body, respectively. For clarity, Video S1 in \cite{Experimental_setup} displays the degree of motion of the flapping body with Line-Guides configuration (c). }}}
    \label{fig:experiment_setup}
\end{figure}
The proposed Natural Hovering Extremum Seeking (NH-ES) framework is implemented using a generic, model-free feedback structure, illustrated schematically in Fig.~\ref{fig:nhes_block}. {Consider a general nonlinear system
\begin{equation}
    \dot{x} = f(x,\boldsymbol{u}),
\end{equation}
where $x$ denotes the system state and $\boldsymbol{u}$ denotes the control input vector. In the experimental flapping-wing system considered here, the control input is composed of two inputs,
\begin{equation}
    \boldsymbol{u} =
    \begin{bmatrix}
        u_1 \\
        u_2
    \end{bmatrix},
\end{equation}
where $u_1$ represents the control input applied to the flapping motor, and $u_2$ represents the control input applied to the pitch servo. Therefore, the system dynamics can be written, as shown in Fig.~\ref{fig:nhes_block}, as
\begin{equation}
    \dot{x} = f(x,u_1,u_2).
\end{equation}}
The function $f(\cdot)$ is assumed to be completely unknown and represents the mechanical system dynamics of {the physical or biological flapping-wing system under consideration}. The objective function $J(x)$ is also assumed to be unknown and may be defined implicitly through sensor measurements, such as light intensity, without requiring an explicit analytical expression. {Each control input is composed of the same slowly varying estimate $\hat{u}$ and a periodic excitation induced by the natural flapping motion, but with different scaling and excitation-amplitude parameters. Specifically,}
\begin{equation}
    {
    u_1 = c_1\,\hat{u} + a_1\omega\cos(\omega t),
    }
\end{equation}
\begin{equation}
    {
    u_2 = c_2\,\hat{u} + a_2\omega\cos(\omega t),
    }
\end{equation}
{where $a_1$ and $c_1$ are the excitation and scaling constants associated with the flapping motor input, $a_2$ and $c_2$ are the corresponding constants associated with the pitch servo input, and $\omega$ denotes the frequency of the oscillatory input.} The adaptive component $\hat{u}$ is updated via induced perturbations resulting from the input signal (flapping motion) on the unknown objective function $J$ measurements: 
\begin{equation}
    \dot{\hat{u}} = k J \omega \cos(\omega t),
\end{equation}
where $k$ is the learning rate. Importantly, no model of $f(\cdot)$ or analytical form of $J(\cdot)$ is required, reflecting that insects/hummingbirds do not need to internalize/comprehend their body/aerodynamic models or their morphological parameters; all that they needed is the oscillatory input naturally generated by the flapping motion itself. 

The experimental validation of NH-ES is conducted using a laboratory-scale flapping-wing body designed to emulate flapping hovering motion. A schematic of the experimental setup and setup configuration is shown in Fig.~\ref{fig:experiment_setup} {while more details and specifics are provided in “Setup.pdf” (also available here \cite{github_flapping_experiment}).} {The flapping body is relatively, but not strictly constrained using parallel lines and two Line-Guides which act as rails to allow motion on the vertical axis but limit lateral displacement. The line tension is relaxed to preserve the natural flapping-induced oscillations. As shown in Fig.~\ref{fig:experiment_setup}, the Line-Guides can be set up in two configurations: far-apart (Fig.\ref{fig:experiment_setup}(b)) which severely limits the body ability to pitch, effectively constraining the dynamics to only one-dimensional vertical translation (isolation of the altitude dynamics), and close-apart (Fig.\ref{fig:experiment_setup}(c)) which allow pitching motion (see Fig.\ref{fig:experiment_setup}(d) and (e)) on top of vertical displacement, hence expanding the experimental setup to two degrees of freedom; for visual clarity, the reader may see the Supplemental Material (Video S1 in \cite{Experimental_setup}). These two configurations give us the ability to run experiments with constrained pitch (as in Section \ref{sec:III}) and with adaptive, controlled pitch dynamics (as in Section \ref{sec:2d_NHES}). These two sets of experiments in Sections \ref{sec:III} and \ref{sec:2d_NHES} provide, respectively, the two experimental validations this work aims at: (1) validation of the initial simulations of purely vertical-motion flapping, findings/observations in \cite{elgohary2025hovering}; and (2) demonstrate the ability of the NH-ES to go beyond vertical-motion flapping and include pitching dynamics (i.e., longitudinal mode).}

{The Wing flapping, generated by a single motor command, $u_1$, and the pitching, generated by a single servo command, $u_2$, can be independently actuated. The reader can refer the Supplemental Material (Video S2 in \cite{Long_wing_pitchdyn}) for an experimental test showing isolated pitch actuation which will be used to control pitch dynamics throughout the experiments in Section \ref{sec:2d_NHES}.}

{In the following subsections, before providing our main results (next sections), we run two experimental verifications aiming at verifying that our setup in terms of flapper, sensor reading, motion capture system are operating effectively. First, we show that the natural oscillation induced by the wing flapping induce perturbation in the objective function measurements, then we proceed with a first validation of NH-ES in a set altitude objective scenario that we know (not model-free) to evaluate the functionality of the setup and the design.}

{\textbf{Remark:} While in the first validation of NH-ES an altitude feedback from the motion capture system (see Fig.\ref{fig:experiment_setup} \#1) is used to compute the altitude-based objective function, this was used for evaluating the design and system functionality by providing the desired height of hovering via the motion capture system. However, in all successive main results in the "moth-like" experiments (Section \ref{sec:III} and \ref{sec:2d_NHES}) the only sensory feedback is the measurement of the body-mounted photo-resistive sensor (see Fig.\ref{fig:experiment_setup} \#6) and, most importantly, no aerodynamic or body model, nor system' states measurement or estimation is used in the feedback loop.}

\subsection{Experimental verification of NH-ES input-output fundamentals: natural flapping motion leads to sensory perturbation}
A central premise of NH-ES is that the {high amplitude, high frequency flapping-wing motion (i.e., input perturbation/oscillation)} naturally induces perturbations/variations in sensory measurements {(i.e, output)} taken on the body, thereby providing the excitation required for extremum search {and adaptation. Furthermore, if the aforementioned premise is verified and holding, this confirms that NH-ES is in alignment with the fundamentals, findings and predictions of ES-VS}. The input-output relation is verified experimentally by fixing the flapping motor command while maintaining a constant relative position with respect to a light source, and observing the light sensor measurements taken on the body. In this test, the motor input is held constant at a Pulse Width Modulation (PWM) command of {$40{,}000$}. As shown in Fig.~\ref{fig:hovering_J}, the measured light-intensity signal {(output)} exhibits clear variations induced solely by the flapping-wing  motion {(input)} {and distinguishable from sensor noise. Additional discussions on sensor noise are addressed in Section \ref{sec:noise}.} These experimentally observed oscillations confirm that the flapping-wing motion itself act not only as a natural probing mechanism, but also results in perturbed sensation that is needed for the feedback as predicted by the NH-ES theory. A {representative} experimental recording is provided in the Supplemental Material (Video~S3 in \cite{Long_S1}). 

\begin{figure*}[t]
    \centering    \includegraphics[width=0.9\linewidth]{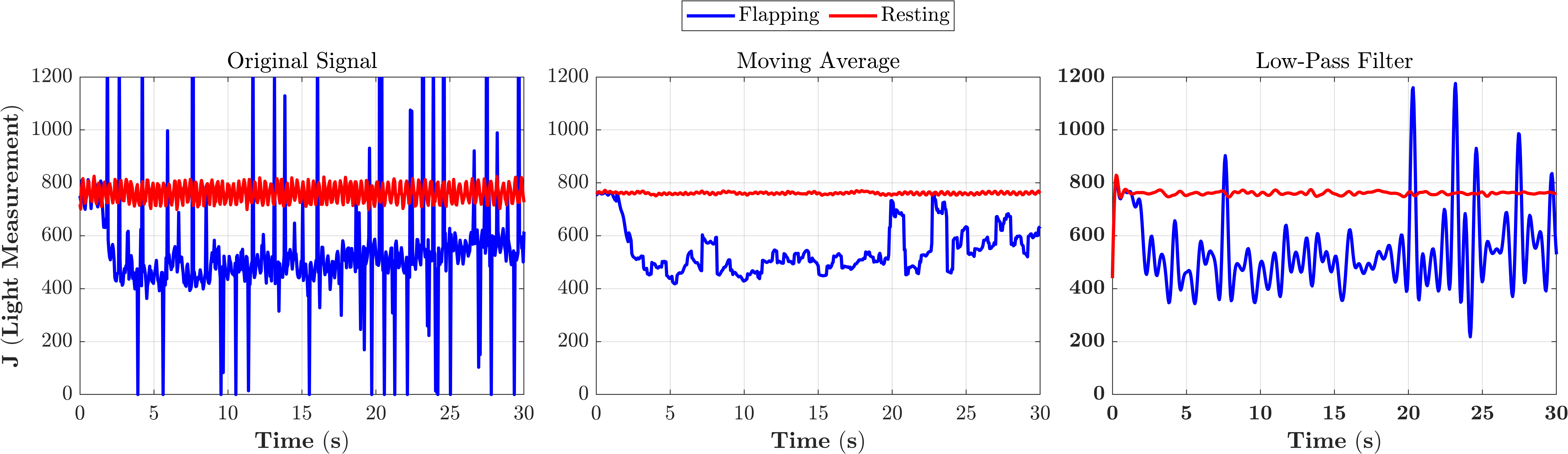}
    \caption{{Natural perturbations in light sensory measurements induced by wing flapping perturbations (blue lines) vs noise-induced light sensory measurements when at rest (red lines). With a constant motor input (PWM $=40{,}000$) and fixed position relative to the light source, light-intensity measurements {(NH-ES output)} exhibit variations/perturbations arising solely from flapping-wing motion {(NH-ES input)}, which is required by NH-ES to adjust flapping wing motion (propulsion) towards stabilization. While the sensor clearly has noise in the measurements, the flapping-induced natural perturbation can be clearly observed. To aid the visualization, from left to right are displayed the original signals, their moving average (20 samples window), and their low-pass filtered plot (cutoff frequency of 1 Hz).}}
    \label{fig:hovering_J}
\end{figure*}

\subsection{Testing the functionality of the design and the system via set altitude-objective}
\label{sec:first_experiment_altitude_objective}

{We first validate NH-ES using a set-altitude objective function, $J(z) = (z - z_d)^2$, defined by the vertical position of the body, $z$, and a desired hovering altitude, $z_d = 700~\mathrm{mm}$ (corresponding to the objective minimum). Since, here, the focus is solely on altitude dynamics, the Line-Guides are set to only allow minimal pitch dynamics (i.e., {Fig.\ref{fig:experiment_setup}(b)} configuration). Starting from rest condition, the flapping-wing body autonomously ascends and converges to the objective minimum, then stable hovering is achieved as shown in Fig.~\ref{fig:fixed_light}. The results demonstrate that hovering stabilization emerges directly from the interaction between intrinsic flapping-induced oscillations and scalar feedback of altitude measurement. A video of this experiment is provided in the Supplemental Material (Video~S4 in \cite{Long_S2})}.

\begin{figure}[ht]
    \centering
    \includegraphics[width=0.9\linewidth]{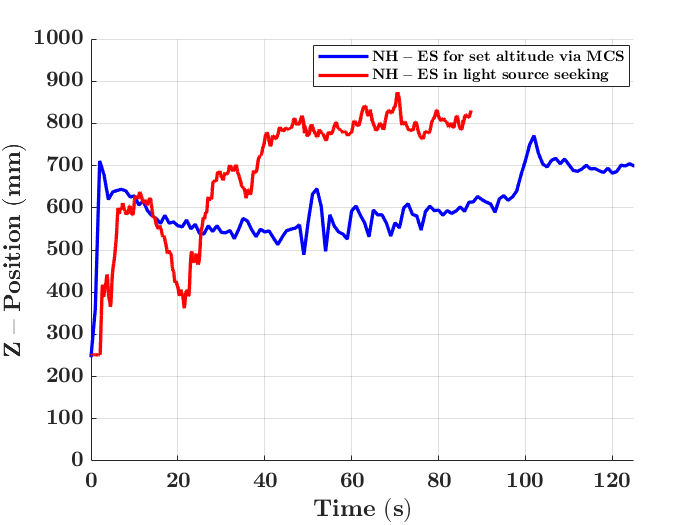}
    \caption{{Blue: experimental validation of NH-ES with set-altitude objective function (NH-ES parameters: $\omega=100$, $k=0.003$, $a=0.7$, $c=1.095$). Red: experimental validation of NH-ES in a completely model-free, moth-like source seeking of unknown light source (NH-ES parameters: $\omega=120$, $k=0.5$, $a=0.7$, $c=1.1$, high-pass filter gain $h=0.2$). In both experiments, the flapping-wing body autonomously ascends and converges to a stable hovering state about the desired objective (set-altitude or maximum light intensity, i.e., light source). 
    The NH-ES control mechanism does not have any access to morphological parameters, light source position or distribution model/estimate, or bode/aerodynamic models.}}\label{fig:fixed_light}
\end{figure}

\subsection{Transition to our main results: completely model-free, sensory-based moth-like experiments}
{Having shown that NH-ES can achieve stable hovering around a set altitude objective in the previous subsection and verifying the flapper, design and motion capture system functionality, in Section \ref{sec:III} and Section \ref{sec:2d_NHES} we proceed in testing its effectiveness in characterizing the biological phenomenon of source seeking (e.g., light source seeking) which is observed in several insects such as moths. In these ``moth-like" experiments we use a light bulb to generate a spatially varying light field and the objective function is obtained solely through instantaneous, local light intensity measurements obtained by the light sensor positioned on top of the flapping body (see Fig. \ref{fig:experiment_setup}(\#6)):
\begin{equation}\label{eq:J_light}
    J = \text{Light-Sensor Measurement}.
\end{equation}
Therefore, no system state measurement/estimation (such as altitude and angles) is needed.}

\section{Experimental demonstration of NH-ES with fixed pitch}\label{sec:III}
{In this section, we initially run experiments focused on the vertical motion with minimal pitch dynamics (i.e., {Fig.\ref{fig:experiment_setup}(b)} configuration). That is, in the NH-ES design schematic in Fig.\ref{fig:nhes_block}, only the flapping motor command (i.e., $u_1$ control input) is adaptive and in effect while the pitch servo command (i.e., $u_2$ control input) is fixed. {By the end of the section however, having shown the experimental efficacy of NH-ES in characterizing stable hovering, subsection \ref{sec:1dSwitchServo} opens up the experimental setup to allow pitching motion (i.e., {Fig.\ref{fig:experiment_setup}(c)} configuration) and allows to pose the question motivating Section \ref{sec:2d_NHES}: can the NH-ES characterize stable hovering flight in a two dimensional longitudinal setting (altitude and pitch)?}

\subsection{First moth-like experiment: source seeking {of unknown} fixed light {bulb}}
{We replicate the NH-ES stable hovering flight of Section \ref{sec:first_experiment_altitude_objective} but considering moth-like light source seeking in completely unknown environment. This is done by introducing a spatially varying light field and replacing position feedback with a body-mounted photo-resistive sensor. In this experiment, the objective function $J$ is not known analytically and is defined by the measured instantaneous, local light intensity provided by the sensor on the body \eqref{eq:J_light}, which serves directly as the scalar objective to be maximized. The spatial distribution of the light field and the location of its extremum {(i.e., the source/bulb)} are completely unknown to the {NH-ES} control mechanism. As shown in Fig.~\ref{fig:fixed_light}, the flapping-wing body ascends autonomously toward increasing light intensity and converges beneath the light source, where it naturally transitions into a stable hovering state. This behavior is repeatable across trials and relies exclusively on local scalar measurements. The corresponding evolution of the light-sensor objective $J$ is shown in Fig.~\ref{fig:fixed_light_J}. A corresponding experimental video is provided in the Supplemental Material (Video~S5 in \cite{Long_S3}).}

\begin{figure}[ht]
    \centering    \includegraphics[width=0.9\linewidth]{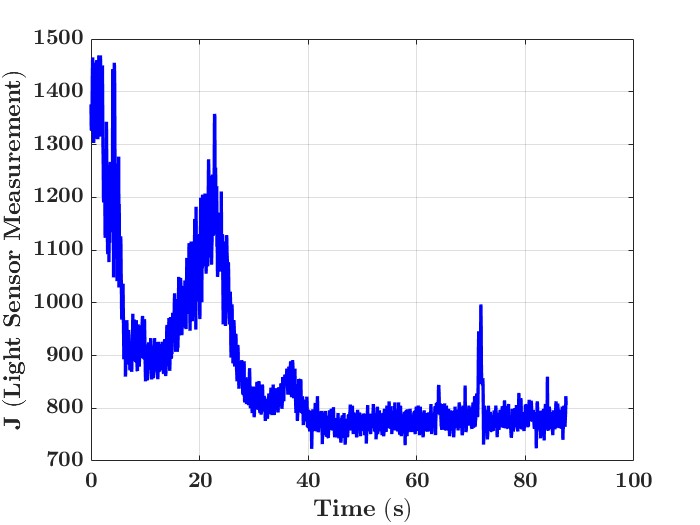}
    \caption{Light-sensor (photoresistive) measurements on the body during the moth-like, light source seeking experiment {(Section III.B)}. The scalar objective $J$, implicitly defined by the measured light intensity, attains its minimum at the location of maximum light intensity, corresponding to a hovering altitude of approximately $z = 800~\mathrm{mm}$, as shown in Fig.~\ref{fig:fixed_light}. {Note that the photosensitive value is inversely proportional to light intensity, so it is minimized at light intensity maximization.}}
    \label{fig:fixed_light_J}
\end{figure}

\subsection{Second moth-like experiment: source seeking {of unknown} moving light {bulb}}
To evaluate adaptability under time-varying conditions, the light source is manually repositioned during flight while keeping the {NH-ES} control mechanism unchanged and without any re-tuning. Despite the moving extremum, the flapping-wing body adapts in real time and re-converges to the new source location, effectively transiting between ascend and hovering as needed, stably. This behavior is illustrated in Fig.~\ref{fig:moving_light}, where the body continuously tracks the changing light source and maintains stable hovering. These results confirm the robustness of NH-ES under dynamic environmental conditions. A video of the moving-source experiment is provided in the Supplemental Material {(Video~S6 in \cite{Long_S4})}.
\begin{figure}[t]
    \centering    \includegraphics[width=1\linewidth]{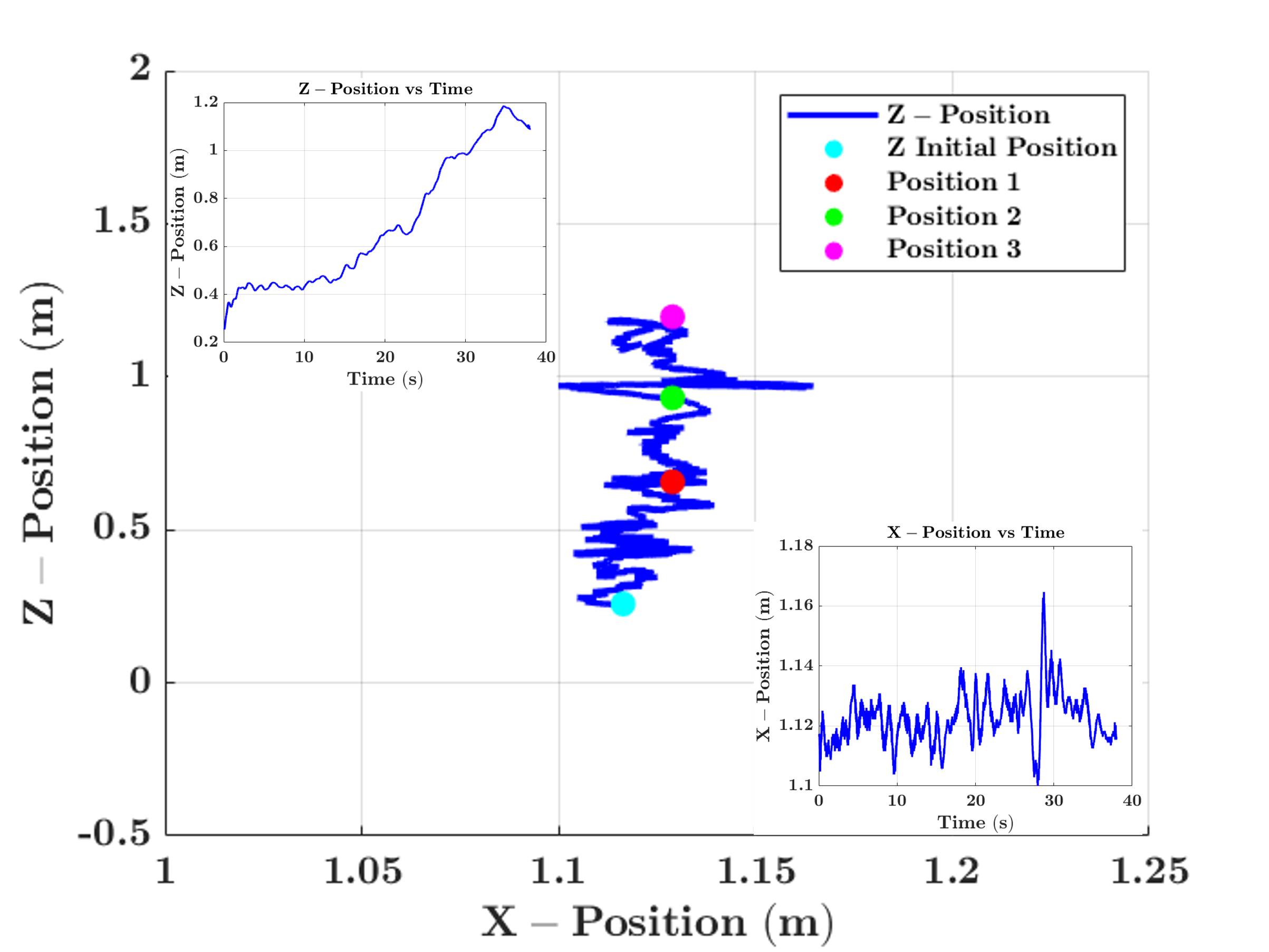}
    \caption{Adaptation to a time-varying extremum (moving light source). The control parameters are identical to those used in the fixed light-source experiment (see Fig.~\ref{fig:fixed_light}). When the light source is repositioned during flight, the flapping-wing body adapts in real time and re-converges to the new source location, demonstrating robustness of NH-ES under dynamic conditions in a complete model-free setting.}
    \label{fig:moving_light}
\end{figure}

\subsection{{Third moth-like experiment: fixed light source seeking with switching servo pitch command}}\label{sec:1dSwitchServo}
\begin{figure*}[t]
    \centering
        \centering        \includegraphics[width=0.9\linewidth]{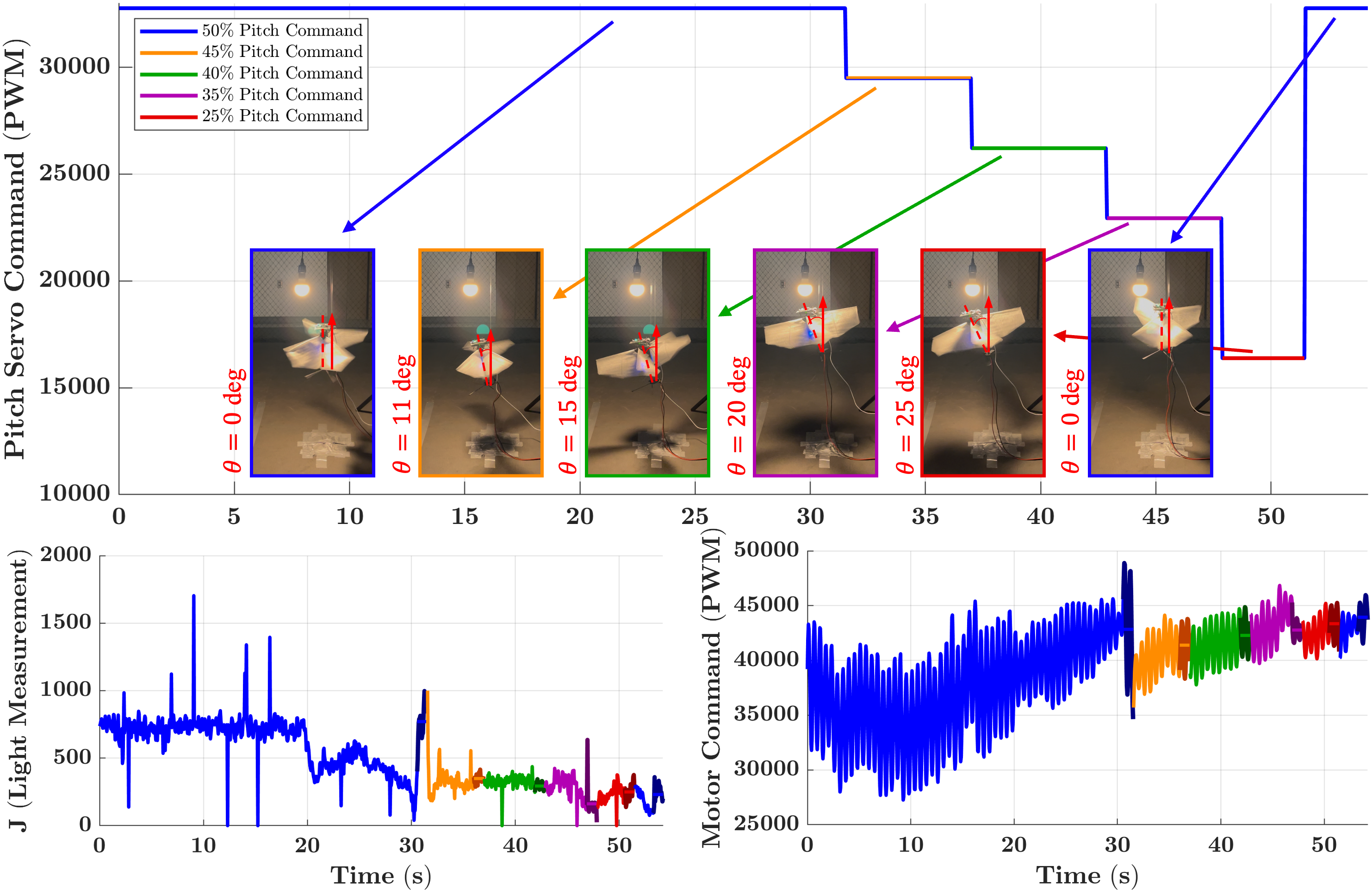}
        \caption{{Experimental results of moth-like, fixed-light source seeking with dynamical transition between several value of fixed pitch servo commands (Section III.C). The experiments uses Fig.\ref{fig:experiment_setup}(c) configuration. In the experiment, NH-ES is obtained through motor command while the pitch servo is provided with several fixed commands to obtain progressively larger pitch angles. The top figure present the pitch servo command evolution over time. Namely the initial pitch servo command of 50\% (blue contour) is switched to 45\% (orange contour), 40\% (green contour), 35\% (purple contour), 25\% (red contour), to then revert back to 50\% (blue contour). The bottom left figure displays the  evolution of light sensor measurement (objective function) while the bottom right graph shows the motor commands. For each pitch command interval, the last 1 s is highlighted (darker shade regions) and the average light measurement and motor command value is displayed. It is clearly observed that different pitch angles (even at similar altitude) correspond to different objective function measurements. Most noticeably, when switching from 35\% (purple contour) to 25\% (red contour), the increase in pitch angle moves the flapping body's light sensor away from the source, resulting in an increase of objective function measurements. Subsequently, when the servo command is set back to 50\%, the objective function reading momentarily drops to then, again, increase. This indicates that, although the flapping robot is able to dynamically stabilize its hovering about different fixed pitch angles, the pitch angle which stabilizes the flapping body around the minimum of the objective function is not automatically obtained using fixed pitch servo commands. Video of the experiment is provided in the Supplemental Material (Video S7 in \cite{Long_switch_fixed_pitch}).}}
    \label{fig:1d_SeveralPitch}
\end{figure*}
{Having shown the the robustness of NH-ES in characterizing stable hovering and flapping flight, even under dynamical environmental conditions in vertical-motion setting, we will transition to testing NH-ES with an additional degree of freedom by changing the experimental setup to Fig.\ref{fig:experiment_setup}(b) configuration, effectively allowing pitching motion throughout the flight. As displayed in Fig. \ref{fig:1d_SeveralPitch}, we investigate if such expansion to allow controlled pitch is possible by experimenting cases where NH-ES is applied only to motor command (like Subsections III.A and III.B) but with several fixed pitch servo commands transitions. Our experimentation was a success. A corresponding experimental video is provided in the Supplemental Material (Video~S7 in \cite{Long_switch_fixed_pitch}). Now that we know we can achieve different fixed pitch commands, and that our experimental setup allows for dynamically changing the pitch, can we control it via NH-ES?} 

{While the NH-ES allows to achieve stable altitude hovering, the pitch angle which stabilizes the flapping body around the minimum of the objective function is not automatically obtained using fixed pitch servo commands. To this end, in the following section, we test NH-ES in a multi-input fashion, effectively adapting the pitch command as well as the motor command.}

\section{Moth-like Experimental Demonstration of NH-ES with Controlled Pitch Dynamics}\label{sec:2d_NHES}
\begin{figure*}[ht]
    \centering
        \centering        \includegraphics[width=0.9\linewidth]{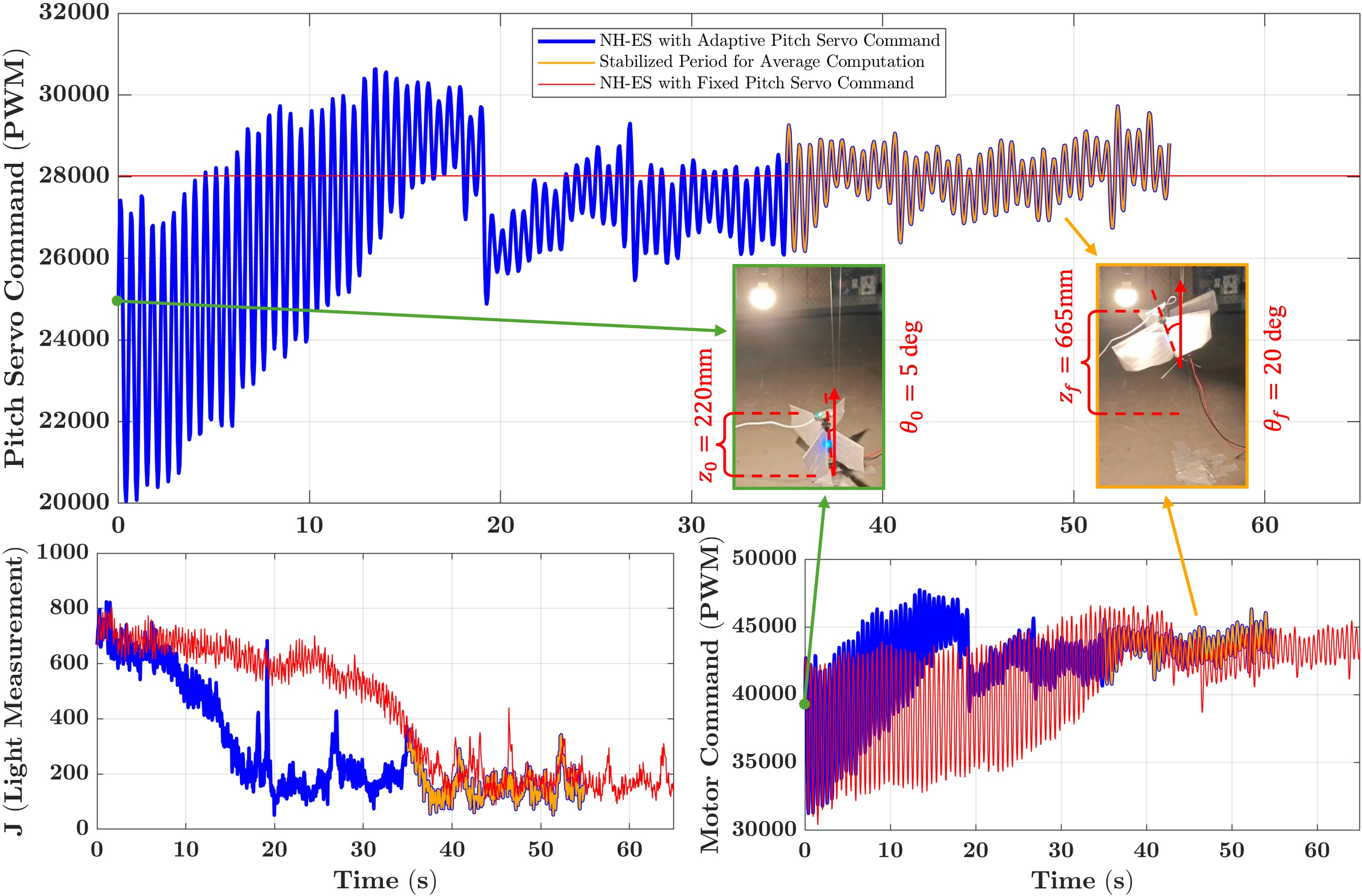}
        \caption{{Experimental results of moth-like, fixed-light source seeking with pitch dynamics (Section IV.A). Two experiments are displayed. First, the 2-D (motor and pitch-servo commands) NH-ES with adaptive pitch servo command (blue line) results in the pitch servo command (top) and motor command (bottom-right) adapting to minimize the light sensor measurement (bottom-left). The first robot figure (green contour) shows the flapper-robot at the start of the experiment, its light sensor (head-part) is at initial height $z_0=220$ mm from the ground and its pitch angle is at around $\theta_0=5$ deg. In the second robot figure (orange contour), the flapper-robot has stabilized to the minimum of the objective function (orange region) with its light sensor (head-part) at about $z_f=665$ mm from the ground (approximately the hight of the lower side of the light source) and its stabilized pitch angle is around $\theta_f=20$ deg. Second, to show that the NH-ES results in a stabilized pitch servo command and motor command which yields a minimum of the light sensor measurement, the average of the pitch servo command is computed for the period in which the flapping robot stabilizes (orange region) and the experiment is repeated in a 1-D fashion (motor command only) by fixing the pitch servo command to said stable average value (red line). It is clear that the NH-ES with fixed stable-averaged servo command results in the motor command to stabilize in practically the same stable region of the 2-D experiment which yields practically the same minimum of the light sensor measurement. It is further verified (but not shown in the interest of space) that the resulting final altitude and angle of the 1-D NH-ES with fixed pitch servo command are approximately identical to the 2-D one. Video of the experiment is provided in the
Supplemental Material {(Video S8 in \cite{Long_fixed_light})}.}}
    \label{fig:main7}
\end{figure*}
{Pitching dynamics (and longitudinal motion) are in the heart of the stability-instability debate in literature, especially for smaller insects \cite{taha2020vibrational,liang2013nonlinear}. Therefore, in this section,
{we run experiments using the experimental setup with Fig.\ref{fig:experiment_setup}(c) configuration  and test the extension of NH-ES to a second degree of freedom.}
That is, in the NH-ES design schematic in Fig.\ref{fig:nhes_block} we have both the flapping motor command (i.e., $u_1$ control input) and the pitch servo command (i.e., $u_2$ control input) adaptive and in effect. It is worth the reminder that NH-ES has not been tested with pitch dynamics in \cite{elgohary2025hovering}. Hence, the results of Section substantially confirm NH-ES premise as a new paradigm in hovering and flapping flight physics.

\subsection{First moth-like experiment with pitch dynamics: source seeking of {unknown} fixed light {bulb}}
{We run an experiment with the same objective as the moth-like, light source seeking in Section III.B. However, we actuate both $u_1$ and $u_2$; that is, we control both the flapping motor and pitching servo via NH-ES. The objective of the experiment then is to show that NH-ES is able to adaptively control the longitudinal motion (vertical and pitching) to have the flapping body ascending towards the light source, and then transitions successfully into a stabilized hovering state, particularly, with stabilized pitching -- see Fig. \ref{fig:main7} (blue). To emphasize more the significance of this result, we averaged out the stabilized pitch servo command (see top-orange in Fig. \ref{fig:main7}), which is about 28000 PMW. Then, we repeated the same experiment successfully using a fixed pitch servo command of about 28000 PMW -- see Fig. \ref{fig:main7} (red). The comprehensive view of the results (the whole of Fig. \ref{fig:main7}) confirms that NH-ES was able to control the pitch adaptively until figuring out the value of about 28000 PMW. Said value, if known beforehand (practically very challenging in unknown environment), could serve as a set value for the pitching command, which NH-ES is able to find autonomously. It is worth to remind the reader that NH-ES was able to achieve model-free, real-time control of both flapping motor and pitch servo using only \textit{local} light measurements with no global or GPS information. A corresponding experimental video is provided in the Supplemental Material {(Video S8 in \cite{Long_fixed_light})}.}

\subsection{Second moth-like experiment with pitch dynamics: source seeking of {unknown} moving light {bulb}}
\begin{figure*}[ht]
    \centering
        \centering
        \includegraphics[width=0.9\linewidth]{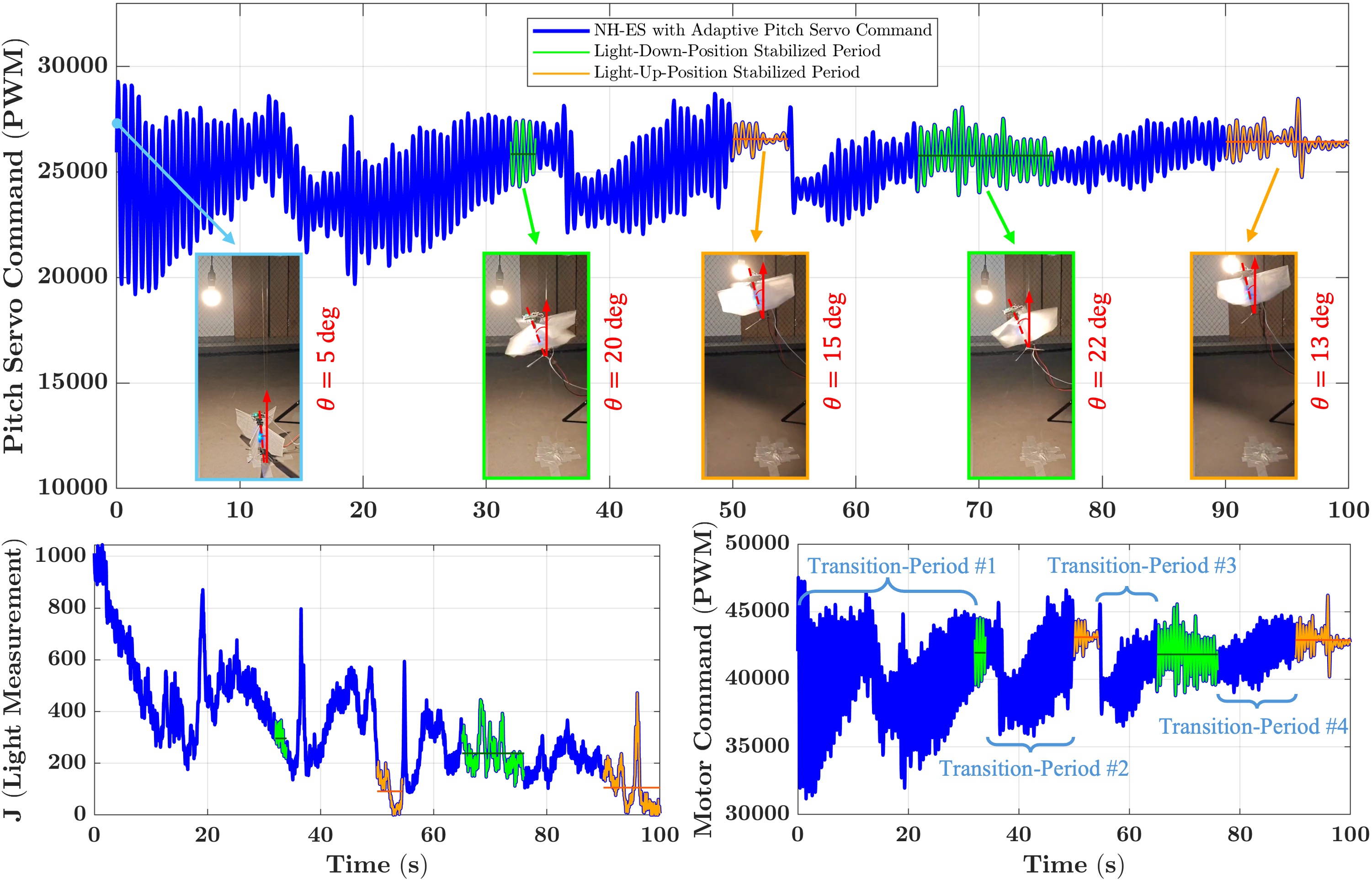}
        \caption{{Experimental results of moth-like, moving-light source seeking with pitch dynamics (Section IV.B). Differently from Fig. \ref{fig:main7}, for each time that the flapping robot stabilizes around a minimum of the objective function (maximum light intensity), the light source is moved. The light is moved in two possible approximate positions: \textit{Light-Down-Position} the light is set at horizontal distance from the constraining lines such that the flapper-robot has to stabilize at higher pitching angle to minimize the objective function. This is the initial position as shown in the left-most robot figure (light-blue contour); \textit{Light-Up-Position} the light is manually brought to a higher and closer-to-the-constraining-line position such that the flapper has to reduce its pitching angle and fly at a higher altitude to minimize the objective function. The robot figures in green and orange contours show the flapper-robot stabilizing to \textit{Light-Down-Position} and \textit{Light-Up-Position} respectively. The NH-ES with adaptive pitch servo command (blue line) consistently results in the pitch servo command (top) and motor command (bottom-right) adapting to minimize the light sensor measurement (bottom-left) despite the light source being moved throughout the experiment. As for the robot figures, the green and orange regions show the stabilized periods and its respective data average in \textit{Light-Down-Position} and \textit{Light-Up-Position}, respectively. It is clearly shown that \textit{Light-Down-Position} and \textit{Light-Up-Position} results in visibly distinct stabilizing average values of light sensor measurement, pitch-servo command, motor command, and, as consequence, pitch angle and altitude.  The stabilized periods are alternating with transition periods as indicated in the motor command figure (bottom-right). From the experiment video in the Supplemental Material {(Video S9 in \cite{Long_moving_light})}, it is clear that, even though the light movements can induce some local overshoots (Transition-Period \#2 and \#3), the NH-ES successfully and consistently re-stabilizes the flapper-robot at an objective function minimum. Moreover, it is clear that in some occasion, such as Transition-Period \#4, that NH-ES can smoothly transition from \textit{Light-Down-Position} to \textit{Light-Up-Position}, effectively highlighting its property of adaptability under time-varying conditions.}}
    \label{fig:main8}
\end{figure*}
{In this experiment, our objective is to test and verify the ability of NH-ES to respond adaptively and effectively to changing light source position with more focus on testing the pitching adaptability. Hence, we conducted a moth-like, source seeking experiment with both $u_1$ and $u_2$ (vertical and pitching) being controlled by NH-ES (similar to Section IV.A). Then, we made changes of the light position between ``up" and ``down" few times and recorded the results -- see Fig. \ref{fig:main8}. The results show the NH-ES ability to respond adaptively to changes, stably transitioning between the changed positions of light in real-time. A corresponding experimental video is provided in the Supplemental Material {(Video S9 in \cite{Long_moving_light})}.}

\section{Discussion and Comments}
\subsection{{Discussion: why NH-ES mechanism stands out as a new simple, biologically plausible paradigm when compared to the simplest literature methods such as PID?}}
{Proportional-Integral-Derivative (PID) controller and its variants (PI and PD) have been suggested and applied as a mean to control hovering and flapping flight in literature -- see for example \cite{Delft_Conference,DelftDissertation2014,feedback25ristroph2010discovering}. In fact, the platform we used in our experimental framework in section III, see also the ``Setup.pdf" supplementary file (also assessable here \cite{github_flapping_experiment}), has the option, while challenging, of using PID control to fly towards a \textit{predefined} reference point/coordinate. Given that PID control is generally considered very simple and low in computational cost, what makes NH-ES stands out and be considered as a categorically different paradigm? We list below some of these immediate and apparent categorical differences.}
\subsubsection{{NH-ES does not need reference signal apriori unlike PID: that is, NH-ES would enable a moth to seek a light source by simply sensing instantaneous, local light intensity without perception of its state or pitch angle and without knowing position information or coordinates of the light source.}}
{A categorical difference between the philosophy of NH-ES and PID mechanisms is that NH-ES operates as a “perturb and observe”, “trial and error”, or “online search and adapt” mechanism with its aim being dynamically optimizing (maximizing/minimizing) an unknown objective function. Thinking about the light source distribution as the objective function as in \eqref{eq:J_light}), this means that NH-ES via instantaneous, local measurements of the light intensity (i.e., the objective function), it will adapt the flapping wing motion (the input) to steer the system with the purpose of maximizing light intensity, which happens about the light source. Hence, NH-ES does not need any predefined/preflight reference, or global information (e.g., access to GPS) about the position or coordinates of the light source (bulb), nor does it need any information or coordinates about the desired hovering point (desired equilibrium point or condition). In fact, the model-free nature of NH-ES mechanism means the flapper does not need to perceive the system states, pitch angle or any parameter at all.}

{On the other hand, PID control mechanisms work based on the philosophy of reducing error between the current state of the system and the desired reference. In other words, the reference needs to be known apriori (preflight) in order for the PID controller to steer the system towards the desired reference. As a result, the model-free, unknown light source (bulb) position experiments in Section III and IV, will not be possible using a direct PID setup since the light source position (reference) is unknown to the flapper and controller.}

{Presumably, flapping insects and hummingbirds (e.g., moth) do not seek a source of light, heat, scent or otherwise, driven by precalculated or predefined position of the source, which at the very least may change based on different environments, but by relying on their sensation to guide them towards the source. This makes NH-ES able to explain and characterize the model-free, sensation-based source-seeking tasks done by flapping insects and hummingbirds in a more biologically plausible mirroring.}

\subsubsection{{NH-ES is naturally built-in and mirrors insects flight dynamics and control structure unlike PID that need external injected input and suffers with complex nonlinear systems}}
{Insect system dynamics and control loop comes down into two phases. First phase is the input phase, through which insects use flapping motion for propulsion, but they continuously need to adapt their flapping motion bias/drift so they fly to the vicinity of the source. Second phase is the output phase, through which sensation feedback is used to then update the input. The NH-ES mechanism characterizes and mirrors exactly the aforementioned loop as can be seen in Fig. \ref{fig:nhes_block}. It is to be noted that NH-ES requires no external input or processing. From control theoretic point view, NH-ES is based on theoretic guarantees that make it applicable directly to time-varying, nonlinear second-order systems \cite{elgohary2025extremum}.}

{However, PID mechanisms cannot be applied directly to insect flapping systems and are not a natural built-in mechanism either. As noted in our discussion in \cite[Section III.C]{elgohary2025hovering} when we studied vertical motion in flapping insects and hummingbirds, and as widely known in control theory, PID methods work ideally on linear autonomous systems and are sub-optimal for nonlinear systems and may struggle significantly in achieving desired performance. They also have to work on autonomous systems and are not directly applicable to time-varying systems like insect flapping mechanics. As a result, using PID mechanisms with flapping systems need extra consideration of operating conditions, averaging analysis in the design stage (not just for stability analysis) to turn the time-varying flapping system into autonomous one, and linearization that may differ from operating condition to another.}

\subsubsection{{NH-ES is superior to optimal PID performance even in a model-based desired reference}}
{In addition to points 1 and 2 above, when we studied vertical motion of flapping insects and hummingbirds \cite[Section III.C]{elgohary2025hovering}, we were able to have some comparisons with PID by considering model-based desired objective altitude (not model-free unknown source since PID needs a reference point unlike NH-ES) \cite[Section III.C]{elgohary2025hovering}, we were able to show that optimal versions of PID differ based on initial conditions and predefined hovering conditions unlike NH-ES. Additionally, we were able to show that universal NH-ES with same parameters outperformed optimal PID in that NH-ES has lesser overshot magnitude (i.e., responds faster in its stabilization to disturbances), lesser settling time (i.e., stabilizes faster through transient states); and has eigenvalues with larger negative real-parts indicating NH-ES has stronger damping property compared to optimal PID -- see \cite[Section IV]{elgohary2025hovering}.}

\subsubsection{{NH-ES is computationally more basic than even PID}}
{One of the most attractive features about NH-ES biologically plausible is its minimal computational complexity. That is, NH-ES is naturally built-in and mirrors insects mechanics, but also requires minimal processing of sensory feedback information. In fact, NH-ES needs only the function of one integrator (one adaptation) while the simplest version of a PID requires three integrators (three adaptations) in addition to other computations to adjust the parameters if they are not fixed or need changes based on operating conditions. As a result, NH-ES is computationally more basic than PID and needs lesser information to operate naturally with flapping systems. We did summarize a comparison between NH-ES and PID, including computational simplicity, in our earlier study on vertical motion flapping insects and humming birds which the reader can reference \cite[Table IX]{elgohary2025hovering}.}

\subsection{Comments on Delays and Noise}\label{sec:noise}
{When NH-ES was first theorized in \cite{elgohary2025hovering}, we tested it with simulations including  delays, representing insects delays in feedback or processing, and noise, representing sensation noise \cite[Section III.D]{elgohary2025hovering}. Such demonstration of NH-ES made its case stronger as it is inevitable to have delays and noise in biological systems.} Similarly, {our experimental implementation (Sections III-IV) naturally includes several non-ideal effects that are unavoidable in physical robotic systems. These effects include sensor noise, wireless communication delay, computation delay, sampling effects, and actuator response delay. In all light-source-seeking experiments, the measured objective-function signal is obtained from the analog light sensor mounted on the flapper. This signal is transmitted from the Arduino Nano to the Python receiver, where the NH-ES control law is implemented, and the resulting motor and pitch-servo commands are then transmitted to the Flapper Nimble+ board. A simplified illustration of this discrete-time communication and control architecture is provided in Fig.~5 of the supplementary file ``Setup.pdf".}

{In addition to the natural delay sources associated with sensing, communication, computation, and actuation, two \textit{explicit} delays were included in the experimental implementation. First, a \(50~\mathrm{ms}\) delay was included in the Arduino light-sensor loop after each sensor measurement and transmission step. Second, a \(0.02~\mathrm{s}\), i.e., \(20~\mathrm{ms}\), delay was included in the Python implementation before transmitting the updated control commands to the flapper board. Therefore, the experiment includes quantified delay effects at both the sensor-measurement side (i.e., the output of NH-ES) and the control-command side (i.e., the input of NH-ES); this is represented by the two rate-transition stages shown in Fig.~5 of the supplementary file ``Setup.pdf".}

{The light sensor measurement also contains noise due to the photoresistive sensing element, ambient-light variations, analog-to-digital conversion, and wireless data transmission. To illustrate this effect, { the red line in Fig.~\ref{fig:hovering_J} shows a \(60\)-second sample of the light sensor measurement for a fixed source/bulb when the flapper is stationary (resting) on the ground.} Since the robot and the light source are not moving during this test, the variations in the measured signal represent the natural sensor noise and measurement fluctuations present in the experimental setup.}

{Despite these practical challenges, the NH-ES controller is implemented directly using the measured objective-function signal. Therefore, the experimental results demonstrate the ability of the proposed model-free feedback law to operate under realistic sensing, communication, actuation, noise, and delay conditions. This is inline with theoretical predictions about NH-ES \cite[Section III.D]{elgohary2025hovering} and its main framework stemming from ES-VS, which in general is able to tolerate significant noise and delays \cite[Section 5]{palanikumar2026model}.}

\section{Conclusive Remark, Implications, and Future Directions}
{This work reported the first experimental demonstration of its kind of \textit{natural hovering extremum seeking} (NH-ES) as a new paradigm and concept in hovering and flapping flight physics. In our moth-like, light source seeking experiments, we were able to demonstrate and verify NH-ES substantial advantages that were predicted in \cite{elgohary2025hovering} and hypothesized as biologically plausible, such as being simple, model-free, real-time, operable by access to only \textit{local} sensation/measurements of the objective function (e.g., local light intensity) without need to any global information (e.g., GPS) or morphological/body/aerodynamic/light-distribution models. That is, the performer of NH-ES needs only to access local sensory information and perform a natural flapping motion that is adaptively updated by the said local sensory measurement in a ``trial-and-error" or ``perturbation-and-observe" fashion. Additionally, NH-ES has been shown to be able to operate under considerable levels of delays and noise successfully. In addition to \cite{elgohary2025hovering}, this study also paved the way for NH-ES potential to address the stability-instability question posed in the literature for open-loop hovering when pitching (longitudinal) motion is in consideration. Experiments in Section IV confirm that NH-ES is able to simultaneously control both vertical and pitching dynamics motion. Hence, we believe NH-ES will have significant implications on the bio-physics and bio-mechanics aspects of hovering and flapping flight. Additionally, the computational simplicity, stability, and easy application of extremum seeking methods, position NH-ES as a new concept with minimal on-board needs that may change preservatives in flapping robotics and Micro-Air-Vehicles (MAVs).}

{In the future, we aim to develop/expand NH-ES theoretically and experimentally to include both longitudinal and lateral motions, reaching 3D source seeking and hovering.}

\bibliography{apssamp}%

\end{document}